\documentclass[sigconf]{acmart}
\usepackage[hide]{chato-notes} %  change show -> hide
\usepackage{listings}
\usepackage{multicol}
\usepackage{tablefootnote}
\AtBeginDocument{%
  \providecommand\BibTeX{{%
    \normalfont B\kern-0.5em{\scshape i\kern-0.25em b}\kern-0.8em\TeX}}}

\setcopyright{rightsretained}
\copyrightyear{2023}
\acmConference{To appear in KDD’23 \vspace{15pt}}
\acmBooktitle{}
\acmDOI{}
\acmISBN{}

\begin{document}
%\fancyhead{}

%%
%% The "title" command has an optional parameter,
%% allowing the author to define a "short title" to be used in page headers.
\title{Fair multilingual vandalism detection system for Wikipedia}

%%
%% The "author" command and its associated commands are used to define
%% the authors and their affiliations.
%% Of note is the shared affiliation of the first two authors, and the
%% "authornote" and "authornotemark" commands
%% used to denote shared contribution to the research.

\author{Mykola Trokhymovych}
\affiliation{%
  \institution{Pompeu Fabra University}
  \city{Barcelona}
  \country{Spain}}
\email{mykola.trokhymovych@upf.edu}

\author{Muniza Aslam}
\affiliation{%
  \institution{Wikimedia Foundation}
  \city{San Francisco}
  \country{USA}}
\email{muniza-ctr@wikimedia.org}

\author{Ai-Jou Chou}
\affiliation{%
  \institution{Wikimedia Foundation}
  \city{San Francisco}
  \country{USA}}
\email{aiko@wikimedia.org}

\author{Ricardo Baeza-Yates}
\affiliation{%
  \institution{EAI, Northeastern University}
   \city{Silicon Valley}
  \country{USA}}
\email{rbaeza@acm.org}

\author{Diego Saez-Trumper}
\affiliation{%
  \institution{Wikimedia Foundation}
  \city{San Francisco}
  \country{USA}}
\email{diego@wikimedia.org}
%%
%% By default, the full list of authors will be used in the page
%% headers. Often, this list is too long, and will overlap
%% other information printed in the page headers. This command allows
%% the author to define a more concise list
%% of authors' names for this purpose.
% \renewcommand{\shortauthors}{Trovato and Tobin, et al.}

%%
%% The abstract is a short summary of the work to be presented in the
%% article.
\begin{abstract}
This paper presents a novel design of the system aimed at supporting the Wikipedia community in addressing vandalism on the platform. To achieve this, we collected a massive dataset of 47 languages, and applied advanced filtering and feature engineering techniques, including multilingual masked language modeling to build the training dataset from human-generated data. The performance of the system was evaluated through comparison with the one used in production in Wikipedia, known as ORES. Our research results in a significant increase in the number of languages covered, making Wikipedia patrolling more efficient to a wider range of communities. Furthermore, our model outperforms ORES, ensuring that the results provided are not only more accurate but also less biased against certain groups of contributors.
\end{abstract}

\maketitle

\section{Introduction}
% \todo[author=Mykola]{
% Do the filtering of the content to meet the number of pages restriction (9 pages)
% }

% The Web is the largest repository of human knowledge after three decades. 
Being the most popular online knowledge source and the eighth most visited website in the world,\footnote{Similarweb Wikipedia.org statistics overview \url{https://www.similarweb.com/website/wikipedia.org}. Accessed 05 December 2022.} Wikipedia has a central role in the Web ecosystem. Its content is frequently used for powering other websites and products, from educational purposes, such as incorporating verified facts into curricula, to artificial intelligence solutions, such as training large language models~\cite{lemmerich2019world,bert}.

Wikipedia is an online encyclopedia that is edited by volunteers. Unlike traditional encyclopedias written by a closed group of experts, Wikipedia relies on the contributions of a large community of users across the globe. This collaborative approach allows a wide range of viewpoints to appear so that information on the site is constantly updated. However, it also requires significant manual work to maintain the site's accuracy and integrity. This includes reviewing and editing new content, monitoring for vandalism, and enforcing community guidelines. %Volunteer editors play an important role in maintaining the quality and accuracy of the information on Wikipedia.

The collaborative nature of Wikipedia is one of its strengths. However, it also presents some challenges. Not all of the content added to the site is of high quality, and it can be difficult to ensure that the information is accurate, true, and/or unbiased. This is partly because many of the site's contributors are inexperienced users who may make mistakes when adding or editing content. Additionally, Wikipedia is also vulnerable to vandalism, which can take the form of harmful edits or the insertion of misinformation~\cite{saez2019online}. 

To address these issues, Wikipedia has a group of dedicated volunteer editors, known as patrollers,\footnote{\url{https://en.wikipedia.org/wiki/Wikipedia:Patrolling}} who work to ensure the accuracy and integrity of the information on the site. These patrollers review and edit articles, monitor for vandalism, and enforce community guidelines. However, their work is not easy, as they have to keep up with the fast pace of Wikipedia, where on average, around 16 pages are edited per second.\footnote{Wikimedia Statistics \url{https://stats.wikimedia.org/}. Accessed 05 December 2022.}

To support the work of patrollers, Wikipedia has developed machine learning (ML) systems such as ORES (Objective Revision Evaluation Service) to assist in identifying potentially damaging changes~\cite{halfaker2020ores}. Even though ORES implements various ML solutions, we focus on the \textit{damaging} model in this work. This model helps patrollers to easily discover harmful contributions, helping to prioritize the content that requires patrolling.  %By assuming that edits that are harmful will be (usually) undone and those that are not harmful will remain, the ORES \textit{damaging} %
%This model is a classifier using the edits and creators' metadata to . Later this signal is used to prioritize the patrollers' work. 

While ML systems can greatly assist patrollers, there are still open problems to be solved. The current system still has limitations regarding language coverage, precision, and bias against anonymous users. The main goal of this work is to create a tool that addresses these open problems.  

The main contributions of this work are:
\begin{itemize}
    \item Introduction of an open-source,\footnote{GitHub repository: \url{https://github.com/trokhymovych/KI_multilingual_training}} multilingual model for content patrolling on Wikipedia, outperforming the state-of-the-art models;
    \item Significantly increasing the number of languages covered by more than 60\%; 
    \item Study the biases of different models and discuss the trade-offs between performance and fairness, 
    \item Model inference productionalization and deployment. % * Deployed and tested the model in production. 
        %\item Collecting multilingual dataset for KI problem
    %\item Content changes feature engineering for modeling
    %\item Building a \textbf{multilingual} system based on content. 
    %\item Analyzing and eliminating biases that discriminate specific cohorts of users.
\end{itemize}

% Therefore, transparency, and explainability of ML-based solutions, are essential to develop a trustworthy AFCS~\cite{smith2020keeping,halfaker2020ores}.

The rest of the paper is organized as follows. Section~\ref{sec_related_work} presents related work, while Section~\ref{sec_system_design} presents the architecture of our system. In Section~\ref{sec_data_preparation}, we explain our data setup giving the evaluation results in Section~\ref{sec_evaluation}. We finish with some conclusions in Section~\ref{sec_conclusions}.
\section{Related Work}
\label{sec_related_work}

In this section, we cover the related work, categorizing it into three main areas: \textit{(i)} Knowledge integrity in Wikipedia; \textit{(ii)} biases and explainability; and \textit{(iii)} language modeling.

\subsection{Knowledge integrity in Wikipedia}
With the rise of Wikipedia, a problem of inappropriate content occurred rapidly after the platform's creation~\cite{10.1007/978-3-540-78646-7_75}. The nature of harmful revisions varies from trolling and hate speech to intended disinformation and even paid promotions \cite{8735592,DBLP:journals/corr/abs-1910-12596,10.1145/3366423.3380055}. 

Initial research papers presented model-based methods for vandalism prevention, defined a problem as a binary classification task, and used generic features like upper case ratio and term frequency~\cite{10.1007/978-3-540-78646-7_75}. Further works also observed the relationship between editing behavior, link structure, and article quality on Wikipedia~\cite{article_behavior}. The study of vandalism detection in other open-source platforms like Wikidata and OpenStreetMap, which observe vandalism patterns and propose detection approaches, also provide valuable insights applicable to our tasks due to shared similarities \cite{heindorf,ijgi1030315}.

Even the first technologies fundamentally transformed the way of editing and administration in Wikipedia \cite{10.1145/1718918.1718941,10.1007/978-3-642-19437-5_23,chin}. More refined works appeared that rethought the problem from different aspects. As a result, the most significant work presents an Objective Revision Evaluation Service (ORES) currently used as a main tool that helps the Wikipedia community evaluate changes to the encyclopedia and assess the article's quality~\cite{halfaker2020ores}. It uses a combination of metadata about the edit and the editor to evaluate each revision, including bad words and informal word features, page, edit, and user metadata. Although nowadays ORES is a key tool that helps the community to improve KI, it has drawbacks that we try to address in our paper: \textit{(i)} limited number of languages support and difficulty to extend it; \textit{(ii)} limited to fixed dictionaries content semantics features; \textit{(iii)} biases against new and not registered users \cite{DBLP:journals/corr/abs-2006-03121,halfaker2020ores}.

The behavior of editors who are not registered on Wikipedia has been controversial~\cite{10.1145/3290605.3300901}. At the same time, the quality control process and the algorithmic tools used to reject contributions are identified previously as significant reasons for the decline in newcomer retention and active users \cite{article_retention}. It shows the importance of analyzing and preserving biases of new algorithms, especially for new and anonymous users.

Our work is inspired by insights and recommendations provided in the paper by Estelle Smith et al., which observes the alignment of model design with the values of communities based on the ORES example \cite{10.1145/3313831.3376783}.

\subsection{Biases and explainability}
Evaluation of content moderation requires not only calculating their accuracy but also taking into account possible biases. For example, model bias against new and anonymous users can cause the active editor's number to decline in long-term perspective \cite{article_retention}.

To evaluate possible user biases, we are using group fairness metrics implemented in AI Fairness 360 open-source toolkit, such as Disparate Impact Ratio (DIR) and Difference in metrics for privileged and unprivileged groups of users~\cite{aif360}.

Moreover, it is important not just to prevent vandalism but also to provide users with an explanation of the decision made by the model. It can be used for further model enhancement and for editors' skills improvement. As for that, we are using tools for local interpretation. They can help to explain every single prediction of a model independently from each other~\cite{SHAP,LIME}.

% SHapley Additive exPlanations (SHAP) and Local Interpretable Model-agnostic Explanations (LIME) \cite{SHAP,LIME}. 

\subsection{Language modeling}
To build a new system for vandalism detection, we want to extract both meta-features of revision and user and semantic features of content changes. Previous works approached it using fixed lists of bad or informal words \cite{halfaker2020ores}. The problem is that it doesn't consider the context where those words were used. Moreover, such lists are not available for all languages, and their creation is costly, making it difficult to cover more languages in Wikipedia. 

At the same time, recent discoveries in the area of Natural Language Processing show the effectiveness of using Masked Language Models (MLMs) for various tasks \cite{bert}. Moreover, it is valuable that one model can be used for multiple languages, because it is easier to maintain and update. In our work, we are using the pretrained \textit{bert-base-multilingual-cased} model and fine-tuning it for our particular task. This model allows working with 104 languages, chosen from the top 100 largest Wikipedias.\footnote{mBERT: \url{https://github.com/google-research/bert/blob/master/multilingual.md}}

% \textit{(i)} bert-base-multilingual-cased, \textit{(ii)} distilbert-base-multilingual-cased, \textit{(iii)} xlm-roberta-base, \textit{(iv)} xlm-align-base \cite{bert,distilbert,roberta,xlmalign}

% https://arxiv.org/pdf/2006.03121.pdf: data collection example, user profile bias
\begin{figure*}[!t]
\centering
\includegraphics[width=\textwidth]{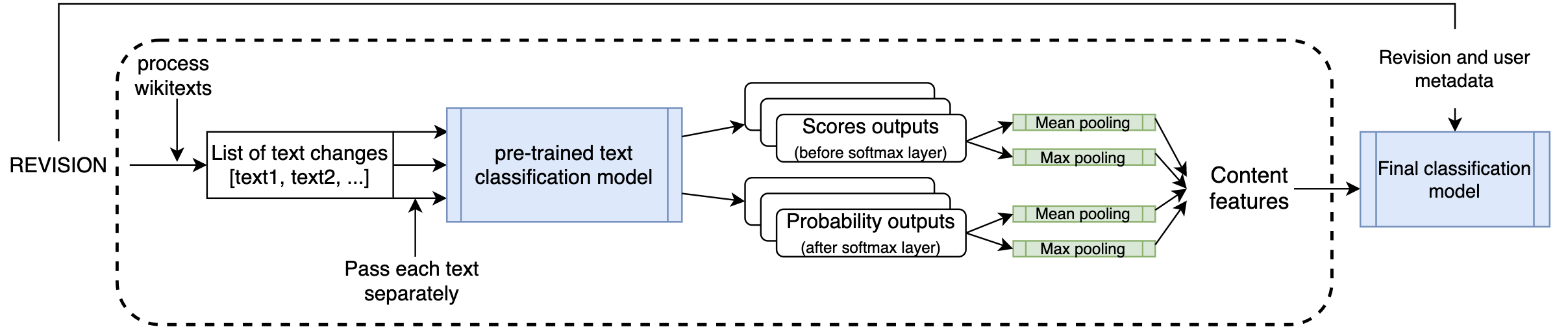}
\caption{System inference logic schema.}
\vspace{-12pt}
\label{fig:screenshot}
\end{figure*}

\section{System design}
\label{sec_system_design}

In this section, we will provide an overview of  the overall architecture and workflow of the system, selecting the appropriate algorithms and technologies, training, testing, and deploying the solution. 

\subsection{Architecture overview}

The proposed system takes the Wikipedia revision as input and aims to define if it will be reverted. It returns the probability of the revert event for a given specific revision.

The key idea of our solution is to analyze the changes in text content, which is the main component of the encyclopedia page. Having the features that represent the content changes, we combine them with the revision and creator metadata to pass in the final classifier to get the revert score.

To sum up, the system pipeline consists of two main steps: \textit{(i)} features preparation; \textit{(ii)} final classification. The system inference logic schema is presented in Figure \ref{fig:screenshot}.

\subsubsection{Features preparation}
\label{content-features}
We start feature preparation with the text-processing steps observed in Section \ref{text-processing}. In the case of the {\tt mwedittypes} package standard output,\footnote{mwedittypes python package: \url{https://github.com/geohci/edit-types}} we use it directly as a feature of the final models. 

As for inserted, removed, and changed texts, we apply the fine-tuned MLMs  that were trained to predict the event of revision to be reverted. Training details will be observed in Section \ref{mlm-training}. Later we use the final model output before (scores) and after (probabilities) the softmax layer as features for the final classifier. We also do the same for page titles that can include valuable semantic information. As each revision can include multiple changes, inserts, or removes, we are calculating features for each and later aggregating them using mean and max pooling. Calculated aggregated scores from each of the four models are later used as input for the final classifier.

\subsubsection{Final classification}
As for the final classifier model, we use the Catboost classification model \cite{catboost}. It is fitted with both text content features observed in the previous subsection and revision metadata. Revision metadata includes the difference in bytes made by revision, users' information, and the language of the page. 

We are training the final model with different feature configurations. The basic configuration includes revisions metadata along with the {\tt mwedittypes} package output. An example of it is presented later in Figure~\ref{actions_example}. Later we are adding MLMs features and user metadata to the basic configuration. The reasoning for separating those features type is that MLMs features require much more computational power to extract, and user features can introduce additional bias to the model. So, we observe the influence of those features on the model performance in different aspects to assess their impact.  

Finally, the classifier returns the probability of the revision being reverted. That score can be further used by patrollers to prioritize work and review the most suspicious revisions at first, constantly improving the quality of content in Wikipedia. 

\subsection{Training process details}
As stated previously, the general system consists of multiple trainable parts related to each other. So, it is crucial to design the training process to avoid any data leakage and finally result in a complex model of high quality. Next, we will observe the process of data splitting and training each part of the complex system. 

\subsubsection{Data splitting logic}
\label{data-splitting}

\begin{figure}[t]
  \centering
  \includegraphics[width=\linewidth]{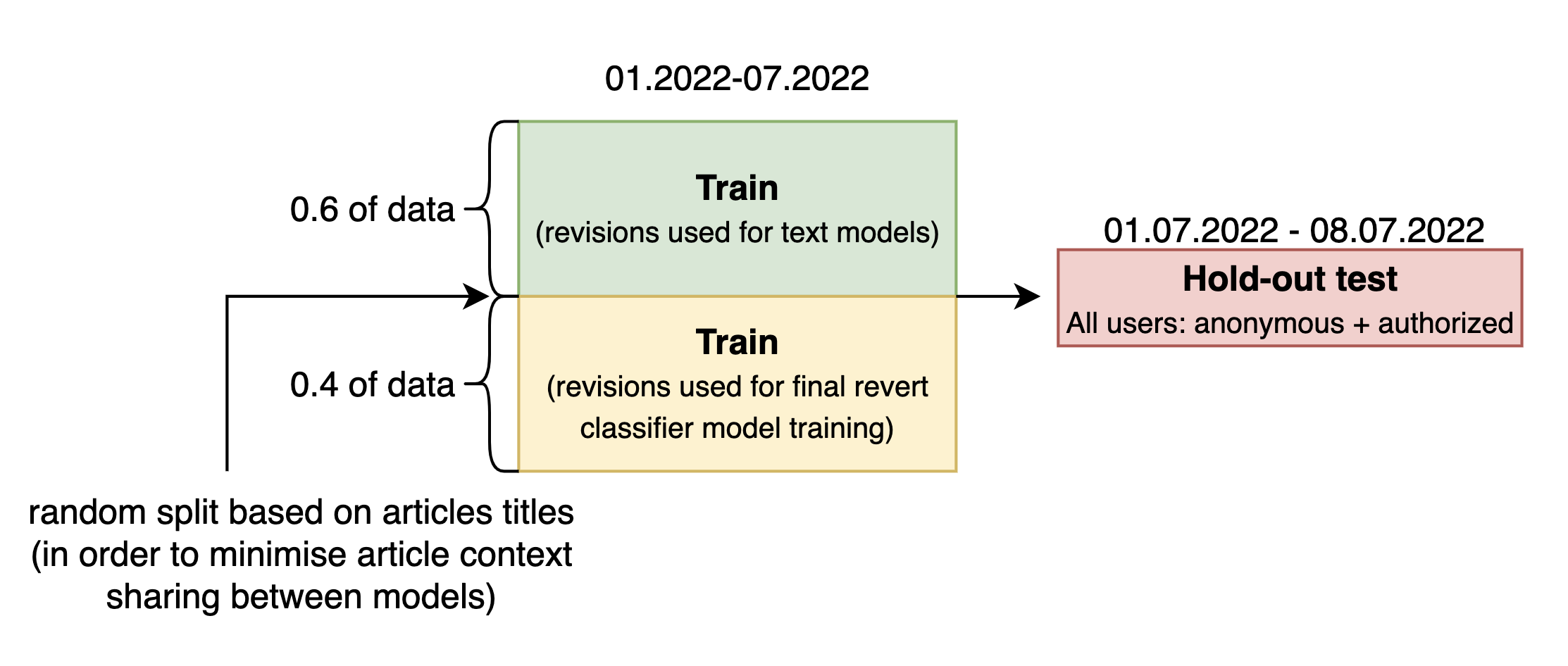}
  \caption{Training-testing split strategy.}
  \label{train-test-split}
  \vspace{-2mm}
\end{figure}

Initially, we have two training datasets of six months of anonymous users and all users' revisions. Also, we have the independent test set of the following week after the training data period. The time-based splitting procedure is needed to avoid time-related anomalies that can bias evaluation results. The overview of the data splitting procedure is presented in Figure~\ref{train-test-split}. This validation data will help to evaluate the final solution in a real-world scenario and compare our system with existing solutions.

As for the training sets, we separated them into two independent parts that were used for fine-tuning MLMs and training the final classifier. We utilized a proportion of six to four, giving more data for fine-tuning MLMs, as they required more data to train, and the procedure observed in section \ref{mlm-training} required heavy filtering. Although splitting was random with the fixed seed, it was secured that revisions of the same articles were not shared between different training datasets. The aim of this condition was to prevent the leakage of article context bias between MLMs and the final classifier. 

\subsubsection{Text classification models}
\label{mlm-training}

The initial approach to modeling was to fine-tune 4 MLMs that will independently evaluate revision features like \textbf{text changes}, \textbf{text inserts}, \textbf{text removes}, and \textbf{pages title semantics}.

Basically, we fine-tune MLMs in three different problem formulations: \textit{(i)} pair of texts classification (text changes), \textit{(ii)} single text classification (text inserts, text removes), \textit{(iii) regression based on text (pages title semantics)}. As for model training, we have resources limited to one AMD Radeon Pro WX 9100 16GB GPU. 

We use the dataset collected for text model training defined in Section \ref{data-splitting}. As for changes and insert model training, we also filter out revisions with more than one change, insert or remove. The aim of this process is to get rid of noise in training data. For example, having two text changes in revision, we can't be sure which of them caused the revision to be reverted. Also, we are doing class balancing for all classification models using undersampling of not reverted revisions.

Data preprocessing for the regression model differs from other models. This model aims to define the revert rate of the page based on the title. For this purpose, we take only those pages that have five or more revisions, whereas other language pages are considered different ones. We calculate the revert rate for those pages and use it as a target for model tuning. 

We use the transformers package for further model tuning \cite{wolf-etal-2020-transformers}. It allows us to load pretrained models directly from the hub and tune them for our specific needs. As for tokenization, we use each model's specific tokenizer. As we have multilingual problem formulation, we use MLM that was initially trained on multilingual datasets - {\tt bert-base-multilingual-cased} \cite{bert}. 
% We use 5\% of data as an internal testset to validate each MLM performance while training. 
During the training process, we use $learning\_rate = 0.00002$, $weight\_decay = 0.01$, and tune models for five epochs. All other parameters are left to be the default. We independently tune models on two different training sets:  train\textsuperscript{anon} and train\textsuperscript{all}, presenting their performance in Section \ref{mlm-metrics}.

\todo{Consider training one model for all the tasks if some time is left}

\subsubsection{Final classifier model}
Having all features for content changes (Section \ref{content-features}) and revision meta-features, we build a classification model to define whether the revision will be reverted. We are using an independent training data subset. With an imbalanced dataset, we apply class weighting proportionally to the ratio between classes. So, we are giving more weight to data samples of underrepresented classes. Also, we have tried out under-sampling of the training data for balancing, but this strategy showed worse results on the final evaluation. As a classification model, we use the CatBoost classifier with $learning\_rate = 0.01$ \cite{catboost}. We train each model for 5 thousand iterations. 

Validation of the model is performed using hold-out testing data presented in Section \ref{data-splitting}, and the final results will be presented in Section \ref{sec_evaluation}. 

% \subsection{Deployment setup and inference}
% \todo{Consider if we should add this section. Ask Aiko and Muniza to help with it.}

\section{Data Preparation}
\label{sec_data_preparation}

% \todo{I suggest to change the order of sections, first you describe the algorithm/system, and then you can go to dataset and explorations}
% The most recent research in the NLP field is highly tacked on to data. Here, we observe and discuss the primary datasets used to solve the NLI problem and their characteristics. 

The modeling stage and further inference performance depend highly on the data used. In this section, we observe the data collection process and explain our technical decisions. Also, we provide the initial exploratory data analysis for the collected dataset. 

\subsection{Data collection overview}

% \todo{Add links to the data when published + update github}

%The data collection logic was performed directly on the Wikimedia Analytics cluster. 
Our data source is the Wikimedia Data Lake,\footnote{Wikimedia Data Lake:~\url{https://wikitech.wikimedia.org/wiki/Analytics/Data_Lake}} specifically,  we used \textit{mediawiki history} and \textit{mediawiki wikitext} history datasets. In particular, the \textit{mediawiki history} dataset stores the metadata about edits to all pages in the form of monthly snapshots. It includes information about the page, the user that performs the change, and details of the individual edit.\footnote{Mediawiki history dumps: \url{https://wikitech.wikimedia.org/wiki/Analytics/Data_Lake/Edits/Mediawiki_history_dumps}} On the other hand, the \textit{wikitext history} contains snapshots of the text content for each of the versions (a.k.a., revisions) made up to the date of the specific snapshot creation.\footnote{Mediawiki wikitext history dumps: \url{https://wikitech.wikimedia.org/wiki/Analytics/Data_Lake/Content/Mediawiki_wikitext_history}}
% \inote{it would be good to briefly describe those tables} 

The collected data have the following characteristics:
% \inote{The filters explained below can be omitted here. Here you can just say the data collected, and below the data that was filtered out}
\begin{itemize}
    \item Contains the 47 most edited languages in Wikipedia to support generalizability. We didn't collect data for Kazakh and Portuguese as they don't allow anonymous revisions, so we could not use them. Also, we omitted \textit{Simple Wikipedia}. 
    \item Contains only records for event entity = 'revision', corresponding to page changes. We limited the number of revisions per language to 300K for training and 100K for testing. There are two main reasons for this restriction: \textit{(i)} we wanted to avoid a huge imbalance in the number of revisions for different languages, \textit{(ii)} we had limited resources for further processing of collected data.  %The logic of the amount limit is implemented using random sampling with fixed random seed.
%    \item Using snapshot dated 2022-07 for data collection.
    \item Uses the period from 2022-01-01 to 2022-07-01 for training and the following week for a hold-out test set.
    % \item Only revisions, which refers to page content changes.
    % \item Revisions created by bots are filtered out.
    \item Includes two training datasets: The first contains only revisions created by anonymous users, and the other contains all users' revisions.
    % \item Revisions created as a part of "revision-wars" are filtered
    % \inote{This requires explanation. RIC: Why this is not mentioned?}
\end{itemize}

In general, the data collection process can be divided into three stages: \textit{(i)} aggregating edits history data; \textit{(ii)} filtering of data explained in the Section~\ref{filtering_section} \textit{(iii)} merging text data and extracting content difference. Text processing details will be explained later in Section~\ref{text-processing}.

The final dataset contains revisions for various languages, their metadata, creator information, and features that correspond to text changes in the revision.

\subsection{Data filtering}
\label{filtering_section}
Wikipedia data is both diverse and noisy. We applied multiple filters to the collected data to construct the proper signal for further modeling. In this section, we describe those filters along with the reasoning behind using them.

\subsubsection{Content type filter}
In this research, we are concentrating on reducing vandalism on content pages. That is why we leave only \textit{'revision'} event entity that corresponds to encyclopedia content page changes. 
% Also, we filter the dataset based on the namespace value. 
Moreover, we omit changes corresponding to discussions, user pages, or other non-encyclopedic content for our research.

It is important to mention that we filter out revisions that create new pages, so they are out of the scope of the proposed solution.  

\subsubsection{User type filter} 
We are removing bots edits from our dataset because we are interested in human-made changes. Also, bot edits present bias towards the lower revert rate and duplicated edits, which can harm model training and evaluation.  

One of the biggest changes differentiating our research from others is leaving only \textbf{anonymous users} edits in the training dataset. Previous works have shown that possible vandalism detection solutions can be biased against anonymous and newly registered users~\cite{halfaker2020ores}. This issue contributes to the decreasing retention rate of newcomers. This is part of an even bigger problem of the declining number of active contributors to Wikipedia~\cite{doi:10.1177/0002764212469365}. Our research contributes to solving this problem that is represented in Section~\ref{fairness_section}.

\subsubsection{"Edit wars" filter}
\begin{figure}[t]
  \centering
  \includegraphics[width=\linewidth]{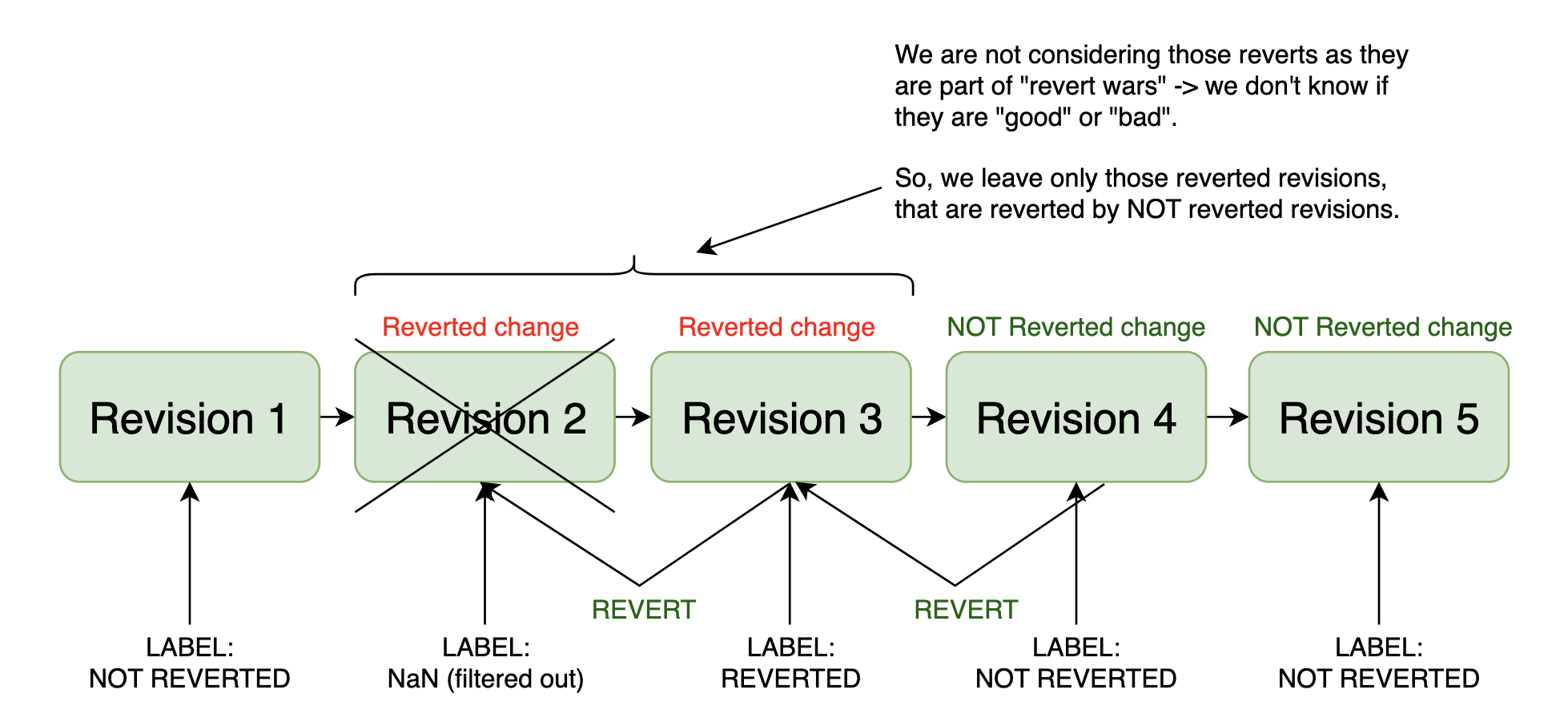}
  \caption{Technique used to filter out "revision-wars."}
  \label{revision-wars}
\end{figure}

Wikipedia article updating is not always a peaceful and collaborative process. There are sometimes strong disagreements in the community. It results in sequential revisions that revert one another, called edit wars. Although those revisions usually include vandalism, we filter them out because it is a source of strong noise in data. The problem is that in the case of edit wars, we have half of the reverted revisions, which are good-faith changes aimed at removing vandalism. At the same time, it is difficult to automatically differentiate vandalism and good-faith changes in the case of edit wars. So we apply a specific filtering technique and remove all reverting revisions that are further reverted by the next revision. The filtering logic is presented in Figure~\ref{revision-wars}.

\subsection{Text processing}
\label{text-processing}
\begin{figure}[t]
  \centering
  \includegraphics[width=\linewidth]{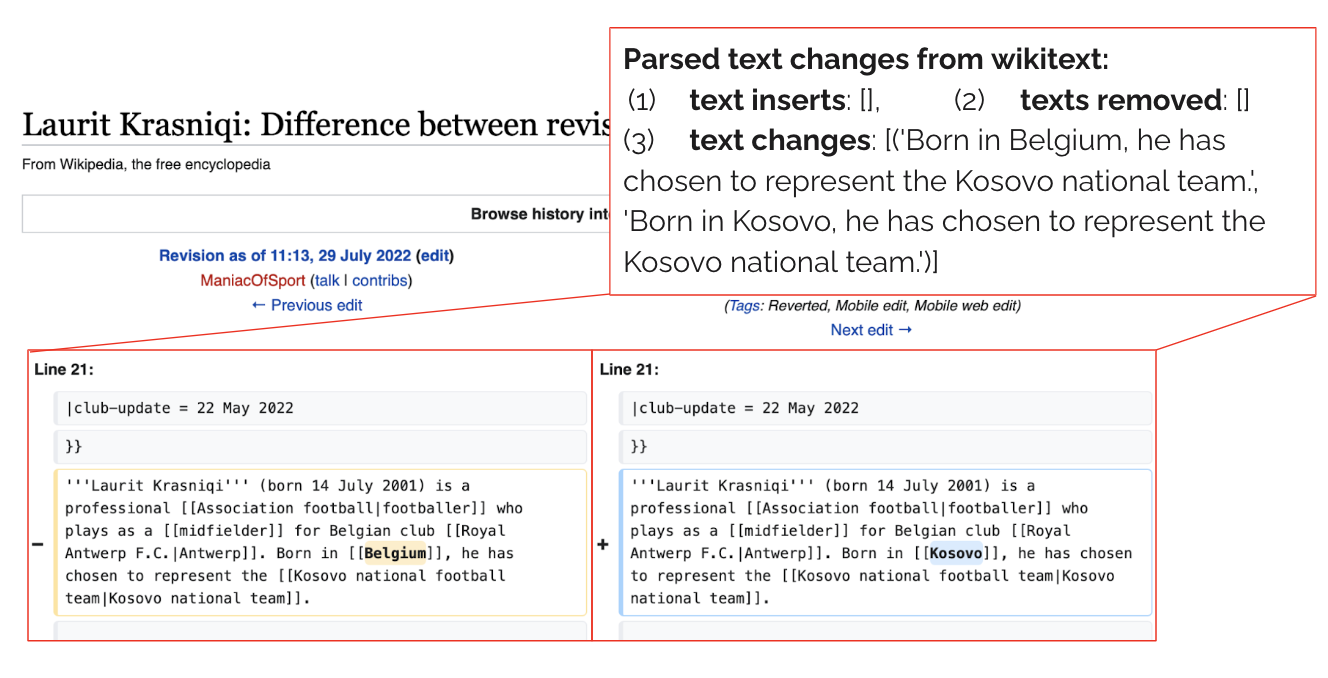}
  \caption{Text content changes extraction.}
  \label{text-parsing}
\end{figure}

Another important step of data preparation is getting features from content changes. Initially, we have two raw strings representing the page's full wikitext content for each revision. We are using the aforementioned {\tt mwedittypes} as the main tool for text change extraction. From a wikitext pair (parent revision and current revision), we extract the \textit{inserts}, \textit{changes}, and \textit{removes} (see Figure~\ref{text-parsing}). Also, we extract the standard output of {\tt mwedittypes} in the form of count for changes, inserts, and removes for different types of data (Text, References, Tables, etc.)

In the case of \textit{inserts} or \textit{removes}, we add all whole parsed texts to the list. They usually represent sentences or phrases. In case of changes, we get the whole paragraph where the change occurred, even if that change is in one character. To extract valuable change signals, we perform the additional step of text processing. We split the extracted paragraph with changes into sentences. Then we compare sentences pairwise using Levenshtein Distance\footnote{
{\tt fuzzywuzzy} python package: \url{https://github.com/seatgeek/fuzzywuzzy}} between the parent and the current revision. 
If there is a minor change in a sentence compared to parent revision, we pass the pair of sentences to the changes list.
If a sentence exists in the parent revision and there is no similar one in the current revision, we add that sentence to the removed list.
If a sentence exists in the current revision and not in the parent one, we add that sentence to the insert list. Figure~\ref{text-parsing} presents an example of text change parsing.

\subsection{Dataset description}
\label{eda_section}

%In this section we provide introductory EDA for the collected data and observe its main characteristics. 
In this section, we provide a general overview of our data. 
We collected three datasets: \textit{(i)} anonymous users training (train\textsuperscript{anon}), \textit{(ii)} all users training (train\textsuperscript{all}), and \textit{(iii)} all users hold-out test set (test). Each of them contains records from 47 languages. The main fields of data schema are presented in Figure~\ref{datasample}. Fields \textit{texts\_removed}, \textit{texts\_insert}, \textit{texts\_change} contain the text content change parsed from wikitext for the specific revision. When \textit{actions} field corresponds to the output of {\tt mwedittypes} and is treated as metadata for the revision. An example of \textit{actions} features is presented in Figure~\ref{actions_example}. 
% \question{What are "actions"? maybe add an example and explanation because it's not intuitive}

\begin{figure}[t]
\begin{center}
\lstset{frame=tb,
  language=Python,
  aboveskip=1mm,
  belowskip=1mm,
  showstringspaces=false,
  columns=flexible,
  basicstyle={\small\ttfamily},
  breaklines=true,
  breakatwhitespace=true,
  tabsize=1
}
\begin{lstlisting}
{'wiki_db': 'enwiki',
  'event_comment': 'grammar',
  'event_user_text_historical': '2603:6010:D800:395B:A03F:E9CD:9BE1:65D7',
  'event_user_seconds_since_previous_revision': nan,
  'revision_id': 1096535207,
  'page_title': 'Balaclava (clothing)',
  'revision_text_bytes_diff': 45,
  'revision_is_identity_reverted': 1,
  'event_timestamp': '2022-07-05 02:46:04.0',
  'revision_parent_id': 1096535065,
  'is_mobile_edit': 0,
  'is_mobile_web_edit': 0,
  'is_visualeditor': 0,
  'is_wikieditor': 1,
  'is_mobile_app_edit': 0,
  'is_android_app_edit': 0,
  'is_ios_app_edit': 0,
  'texts_removed': '[]',
  'texts_insert': '["So, people don\'t have to see your ugly face"]',
  'texts_change': '[]',
  'actions': "{...}",
  'is_anonymous': 1 }
\end{lstlisting}
\caption{Data sample: this revision contains vandalism on the inserted text.}
\label{datasample}
\end{center}
\end{figure}

\begin{figure}[t]
\begin{center}
\lstset{frame=tb,
  language=Python,
  aboveskip=1mm,
  belowskip=1mm,
  showstringspaces=false,
  columns=flexible,
  basicstyle={\small\ttfamily},
  breaklines=true,
  breakatwhitespace=true,
  tabsize=1
}
\begin{lstlisting}
{'change_Media': 0, 'insert_Media': 0, 'move_Media': 0, 'remove_Media': 0, 'change_Punctuation': 0, 'insert_Punctuation': 3, 'move_Punctuation': 0, 'remove_Punctuation': 0, ..., 'change_Whitespace': 0, 'insert_Whitespace': 9, 'move_Whitespace': 0, 'remove_Whitespace': 0, 'change_Word': 0, 'insert_Word': 9, 'move_Word': 0, 'remove_Word': 0 }
\end{lstlisting}
\caption{Example of \textit{actions} features.}
\label{actions_example}
\end{center}
\end{figure}

\begin{figure}[t]
  \centering
  \includegraphics[width=\linewidth]{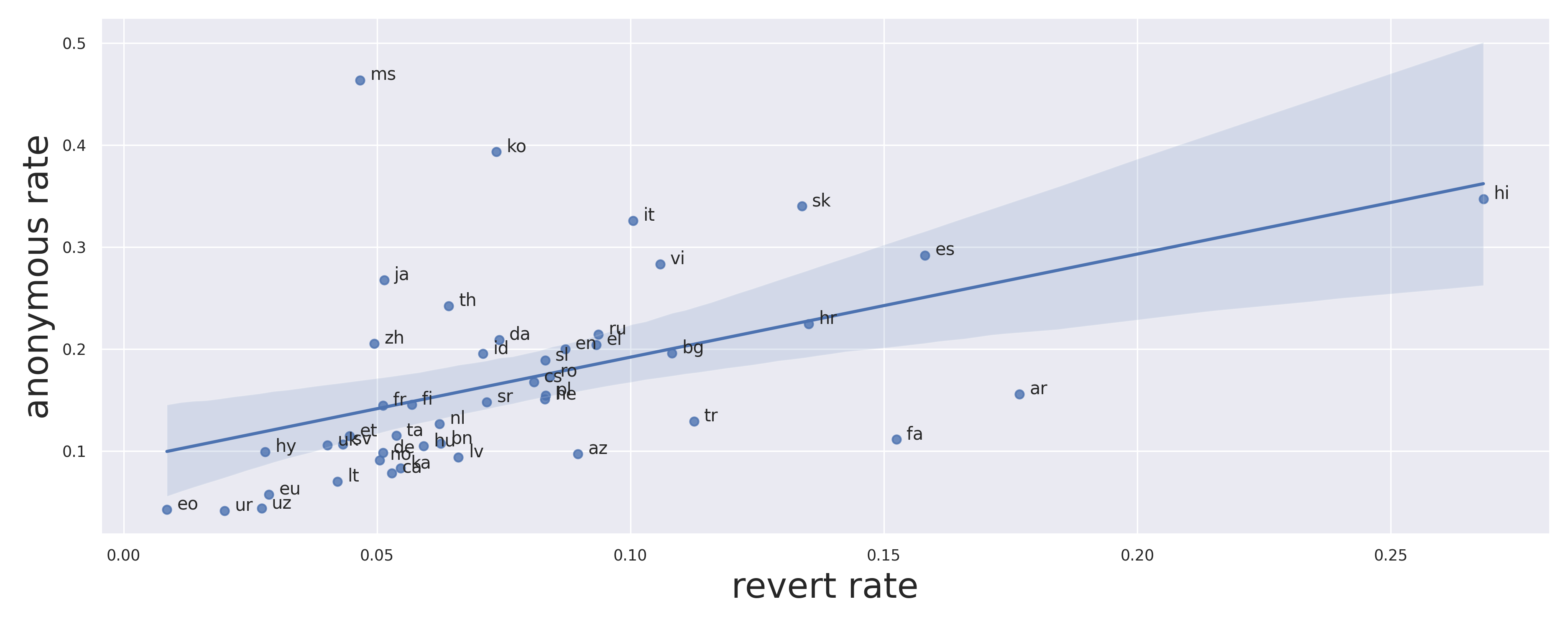}
  \caption{Revert rate vs. anonymous users rate across languages.}
  \label{rates_img}
\end{figure}

Fundamental data characteristics are presented in Table~\ref{data-stats}. Both training sets represent the same period of 6 months. We observe a slightly higher rate of anonymous users in testset. Also, that number differs across the languages. For example, anonymous users' rates vary from 0.04 for urwiki/eowiki/uzwiki to 0.39 for kowiki and 0.46 for mswiki in train\textsuperscript{all}.
As for the revert rate, we observe a significantly higher value in train\textsuperscript{anon} than in train\textsuperscript{all}, as expected. This value deviates from 0.01 for eowiki and 0.02 for urwiki to 0.18 for arwiki and 0.27 for hiwiki for train\textsuperscript{all}. We can observe the relation between the revert rate and anonymous users rate across the languages in train\textsuperscript{all} in Figure~\ref{rates_img}. We also calculated the Spearman correlation coefficient which is 0.62, which supports our reasoning that the anonymous user rate relates to the revert rate.

\begin{table}[t]
\begin{center}
\caption{Data characteristics.}
{\tabcolsep=3pt
\begin{tabular}{p{3cm}|c|c|c}
\hline\label{data-stats}
\textbf{Dataset} & \textbf{train\textsuperscript{anon}} & \textbf{train\textsuperscript{all}} & \textbf{test} \\
\hline
Number of samples & 3,693,571 & 8,586,362 &  1,079,265\\
\hline
Observation period & 6 months & 6 months & 1 week   \\
\hline
Anonymous rate & 1.0 & 0.17 & 0.19  \\
\hline
Revert rate  & 0.28 & 0.08 & 0.07   \\
\hline
\end{tabular}%
}
\end{center}
\end{table}

% \todo{Calculate correlation to support the idea of plotting the line in the plot (in case the correlation is low, than it is not correct representation)}

% \begin{table}[t]
%   \caption{Frequency of Special Characters}
%   \label{tab:freq}
%   \begin{tabular}{ccl}
%     \toprule
%     Non-English or Math&Frequency&Comments\\
%     \midrule
%     \O & 1 in 1,000& For Swedish names\\
%     $\pi$ & 1 in 5& Common in math\\
%     \$ & 4 in 5 & Used in business\\
%     $\Psi^2_1$ & 1 in 40,000& Unexplained usage\\
%   \bottomrule
% \end{tabular}
% \end{table}

% \begin{figure}[t]
%   \centering
%   \includegraphics[width=\linewidth]{sample-franklin}
%   \caption{1907 Franklin Model D roadster. Photograph by Harris \&
%     Ewing, Inc. [Public domain], via Wikimedia
%     Commons. (\url{https://goo.gl/VLCRBB}).}
%   \Description{A woman and a girl in white dresses sit in an open car.}
% \end{figure}
\section{Evaluation}
\label{sec_evaluation}

In this section, we observe the performance of the proposed solution and compare it to other existing models. The goal of our models is to predict whether a given revision will be reverted or not. We consider the accuracy, efficiency, and fairness of models. Moreover, we evaluate our model trained on data with different configurations: \textit{(i)} using different feature sets; \textit{(ii)} training only on anonymous revisions, or using both registered and anonymous users' revisions as training data. 

\subsection{Masked Language Models metrics}
\label{mlm-metrics}
In this section, we evaluate the standalone performance of fine-tuned MLMs for the task of revert prediction. In particular, we evaluate the models trained to predict reverts based on changes, removes, and inserts of the text to the content of the pages. We should mention that using those models in standalone mode is inappropriate, as they will not cover all the revisions. The reason is that it is possible that revisions don't have any text changes, and the difference is only in media, references, or other content. In particular, for our hold-out test set, the rate of revisions with text changes is 38.4\%, text inserts are 24.3\%, and text removes 14.4\%. In general, we have the situation that 55.6\% of revisions have at least one of the text manipulation types. However, in this section, we want to show that MLMs can provide a powerful signal that can be used side-by-side with meta-features, improving the general system performance that is shown in the next Section \ref{final-validation}. The results are presented in Table~\ref{table_mlm}.

We calculate the AUC metric separately for all users (AUC\textsuperscript{all}) and for anonymous user (AUC\textsuperscript{anon}) revisions. We use the aggregation logic observed previously in case of multiple text modifications for one revision. In particular, we are using mean aggregation of probability scores obtained from MLMs. We are only considering revisions having a specific text manipulation used for model training. We use \textit{bert-base-multilingual-cased} as a base model and fine-tune it using all or anonymous only users' revisions. For models fine-tuned on different user group revisions, we present anonymous and all users metrics. 

The first outcome of the analysis is that the model learns the signal from different text manipulations, thus, can be used to improve the complete anti-vandalism system. Another insight is that models fitted on anonymous or all users' revisions perform better on specific types of revisions that were used for training. This leads us to conclude that anonymous revisions have specific patterns of text changes that differ from general cases. Those models' output will be used later for building the final classifier.

\begin{table}[t]
\begin{center}
\caption{Standalone MLM performance for changes, inserts and removes on the test set.}
\label{table_mlm}
{\tabcolsep=3pt
\begin{tabular}{p{2.3cm}|p{2.3cm}|p{2.3cm}}
\hline
\textbf{Model} & \textbf{AUC\textsuperscript{all}} & \textbf{AUC \textsuperscript{anon}}\\
\hline
changes\textsuperscript{all} & \textbf{0.745} & 0.732  \\
\hline
changes\textsuperscript{anonymous} & 0.725 &  \textbf{0.757} \\
\hline
\hline
inserts\textsuperscript{all} & \textbf{0.742} & 0.724  \\
\hline
inserts\textsuperscript{anonymous} & 0.716 & \textbf{0.752}  \\
\hline
\hline
removes\textsuperscript{all} & \textbf{0.662} & 0.654  \\
\hline
removes\textsuperscript{anonymous} & 0.625 & \textbf{0.681} \\
\hline
\end{tabular}%
}
\end{center}
\end{table}

\subsection{Final model}
\label{final-validation}

The following section presents an examination of the performance of the complete system. The results are analyzed and discussed in detail to provide insight into the performance of the model and its potential applications. We compare our model with the existing SOTA solution - ORES. Also, we use a baseline, a biased rule-based model that reverts all the anonymous revisions. 

% We use the AUC score on a full unbalanced holdout test set as the main metric for model comparison. Additionally, we calculate Accuracy and F1 scores with a default threshold (0.5) on the balanced holdout test set. We are balancing by downsampling the overrepresented class across each language.

We use the AUC score on a full unbalanced holdout test set as the primary metric for model comparison. It can be interpreted as a probability of the model ranking a random positive example higher than a random negative example, which directly corresponds to our goal.

% As an alternative metric we use is precision at a recall level of 0.75 ($Pr@R0.75$), which is business oriented. The reasoning is that patrollers can't always review all revisions in real-life usage scenarios, so they select the most suspicious ones. At the same time, various models can have different optimal thresholds, and they are usually selected based on the patrollers' workload for different communities. That is why precision at a recall level is a sub-optimal approach to comparing models' performance in a practical setup. We are calculating this metric using an unbalanced dataset, the same as for the AUC score. Moreover, we are calculating the same metrics for anonymous users only.

% new paragraph
As an alternative metric, we use precision at a recall level of 0.75 ($Pr@R0.75$) on an unbalanced dataset. The reasoning is that patrollers can't always review all revisions in real-life usage scenarios, so they select the most suspicious ones. At the same time, various models can have different optimal thresholds, and they are usually selected based on the patrollers' workload for different communities. That is why precision at a recall level is a sub-optimal approach to comparing models' performance in a practical setup.

We have also calculated MacroF1 on unbalanced along with Accuracy and F1 on the balanced holdout test set. We use a default threshold (0.5) and balance by downsampling the overrepresented class across each language.  With this threshold, our best configuration shows a performance of 4.5\% worse than ORES on all users testing using MacroF1 score. However, those metrics depend highly on the classification threshold, which is inadequate for model comparison in our setup.

The full results of the models' evaluation are presented in Table \ref{table_final_all} for all users and Table \ref{table_final_anon} for anonymous-only users. Only revisions for languages implemented in ORES are used in the presented comparison tables. 

\begin{table}[t]
\begin{center}
\caption{System performance on test set of all users.}
{\tabcolsep=3pt
\begin{tabular}{p{2.2cm}|p{1.2cm}|p{1.4cm}|p{1.2cm}|p{1.3cm}}
\hline\label{table_final_all}
\textbf{Model} & \textbf{AUC} & \textbf{Pr@R0.75} & \textbf{F1} & \textbf{Accuracy}\\
\hline
Rule-based & 0.75 & 0.07 & 0.72 & 0.74  \\
\hline
ORES &  0.84 & 0.22 & 0.65 & 0.71 \\
\hline
Multilingual\textsuperscript{anon}& 0.77 & 0.14 & 0.65 & 0.66  \\
\hline
Multilingual\textsuperscript{anon} + MLM features & 0.79 & 0.15 & 0.66 & 0.68  \\
\hline
Multilingual\textsuperscript{all}  & 0.82 & 0.18 & 0.72 & 0.71  \\
\hline
Multilingual\textsuperscript{all} + MLM features  & 0.84 & 0.20 & 0.73 & 0.73  \\
\hline
Multilingual\textsuperscript{all} + user features  & 0.87 & 0.27 & 0.78 & 0.78  \\
\hline
Multilingual\textsuperscript{all} + user features + MLM features  & \textbf{0.88} & \textbf{0.28} & \textbf{0.79} & \textbf{0.78}  \\
\hline
\end{tabular}%
}
\end{center}
\vspace{-6mm}

\end{table}

We are comparing models trained on anonymous users' revisions only and on revisions of all users. Also, we experiment with adding user-specific features, like user group and is\_anonymous, and MLMs features. It is important to mention that the MLM used for the anonymous user training set was fine-tuned only on anonymous users' revisions. Similarly, all users' training data uses MLMs tuned on all user's revisions. 

\begin{figure}[t]
  \centering
  \includegraphics[width=\linewidth]{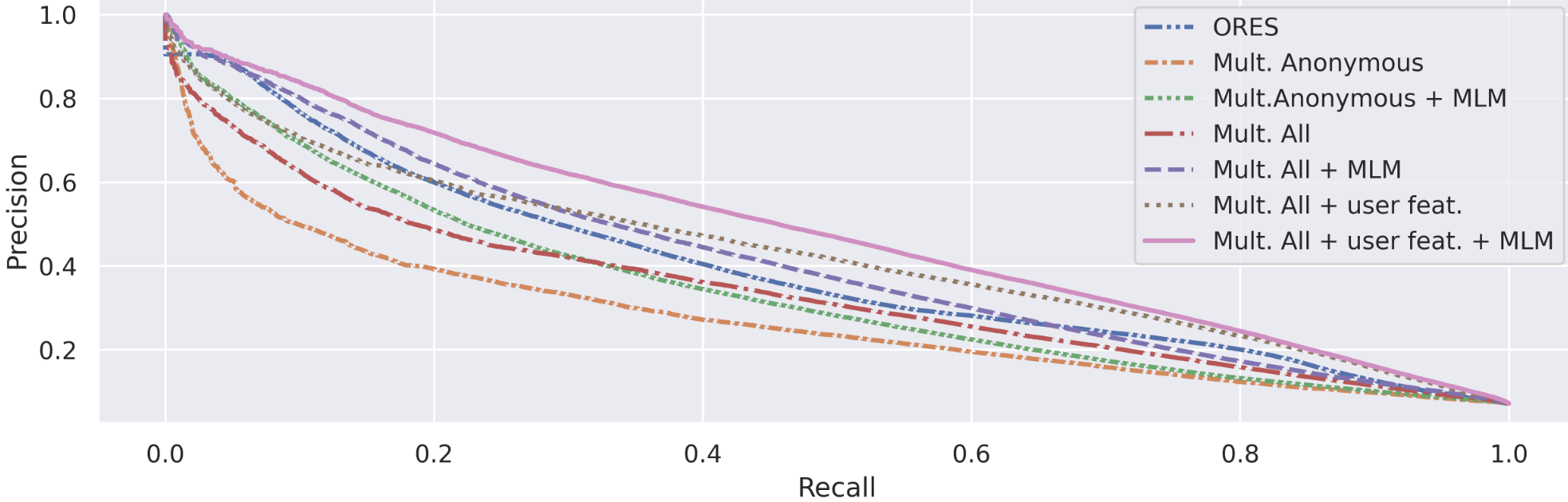}
  \caption{Precision-recall graph for all users.}
  \label{pr_plot_all}
\vspace{-5pt}
\end{figure}

\begin{figure}[t]
  \centering
  \includegraphics[width=\linewidth]{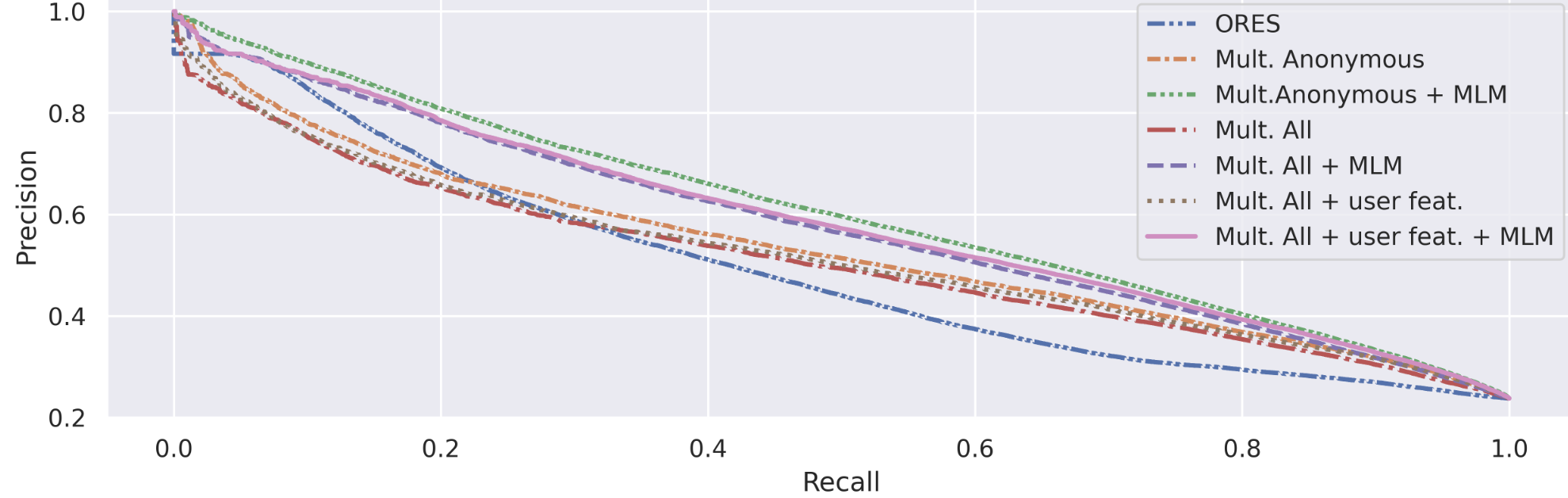}
  \caption{Precision-recall graph for anonymous users.}
  \label{pr_plot_anon}
\vspace{-10pt}
\end{figure}

Finally, we present the precision-recall graphs in Figure~\ref{pr_plot_all} and Figure~\ref{pr_plot_anon} that show how the precision and recall metrics differ for different thresholds, which can be useful for selecting the model for specific community conditions. For example, if the patrollers have significantly limited resources, they will select only the most suspicious revisions to check. It is clear that recall, in that case, will not be high, but at least we want to maximize the precision to spend the patrollers' time efficiently. 

The final results show that the multilingual model trained on all users' revisions with users and MLMs features outperforms all competitors, including ORES, using the AUC metric. Figure~\ref{pr_plot_anon} supports our reasoning and shows that the mentioned model performs the best for any classification threshold. Moreover, we also observe a significant improvement in performance for anonymous users. In particular, we see that MLMs features contribute to all model configurations, especially for anonymous users. 

Furthermore, we evaluate the performance of the best-performing model for languages that are not implemented in ORES. The AUC value for those languages is 0.873, which is similar to those used in the comparison table.

Another important observation is that models trained on anonymous user revisions can only be beneficial for the classification of anonymous revisions. However, they have significantly worse performance considering all users' revisions. Even though the best-performing model for anonymous users appears to be the one fitted with only anonymous users' revisions, we observe that the configuration performing the best for all users has comparable results also on anonymous users. As a result, we select the configuration trained on all revisions with user and MLMs features as the best model and use it for more detailed analysis in the next sections.

\begin{table}[t]
\begin{center}
\caption{System performance on test set of anonymous users.}
{\tabcolsep=3pt
\begin{tabular}{p{2.2cm}|p{1.2cm}|p{1.4cm}|p{1.2cm}|p{1.3cm}}
\hline\label{table_final_anon}
\textbf{Model} & \textbf{AUC} & \textbf{Pr@R0.75} & \textbf{F1} & \textbf{Accuracy}\\
\hline
Rule-based &  0.50 & 0.24 & 0.67 & 0.50  \\
\hline
ORES & 0.70 & 0.31 & 0.63 & 0.57 \\
\hline
Multilingual\textsuperscript{anon}&  0.77 & 0.40 & 0.65 & 0.63  \\
\hline
Multilingual\textsuperscript{anon} + MLM features & \textbf{0.80} & \textbf{0.44} & 0.68 & \textbf{0.67} \\
\hline
Multilingual\textsuperscript{all}  &  0.75 & 0.38 & 0.68 & 0.61  \\
\hline
Multilingual\textsuperscript{all} + MLM features  & 0.78 & 0.42 & \textbf{0.70} & 0.64  \\
\hline
Multilingual\textsuperscript{all} + user features  & 0.76 & 0.39 & 0.67 & 0.53  \\
\hline
Multilingual\textsuperscript{all} + user features + MLM features  & 0.79 & 0.43 & 0.68 & 0.55  \\
\end{tabular}%
}
\end{center}
\end{table}

\todo{Observe how model results coincide (same errors or not)}

\subsection{Model productionalization}

In this section, we analyze deployment details and the efficiency of the complete system. It includes data collection using the MediaWiki API,\footnote{MediaWiki API \url{https://www.mediawiki.org/wiki/API:Main_page}} data processing, feature engineering, and final model prediction. The whole model inference pipeline is open-source and can be found in the repository.\footnote{\url{https://gitlab.wikimedia.org/repos/research/knowledge_integrity}} Service was experimentally deployed using Lift Wing infrastructure.\footnote{Lift Wing \url{https://wikitech.wikimedia.org/wiki/Machine_Learning/LiftWing}} Lift Wing is a set of production Kubernetes clusters at Wikimedia based on KServe,\footnote{KServe~\url{https://github.com/kserve/kserve}} a serverless inference platform. Lift Wing hosts machine learning models and responds to requests for predictions. Machine learning models are hosted on Lift Wing as Inference Services, which are asynchronous microservices and containerized applications. Each inference service has production images that are published in the WMF Docker Registry.\footnote{WMF Docker Registry \url{https://docker-registry.wikimedia.org/}} The final model was deployed to the ML staging cluster and allocated up to 4 CPUs and 6 Gi memory resources to the pod. The service endpoint is exposed using Istio as an ingress to the API consumers.

As for the efficiency experiment, we select a random sample of 1K revisions and pass them sequentially to the Lift Wing API. We use the model with the best configuration for the test. We used the model that requires both revision metadata and MLM-based features. The inference is tested on CPU-only instances. As a result, we got the median response time of 1.0 seconds, 1.7, and 4.0 seconds for 75\%, and 95\% percentile, respectively. It should be mentioned that revision processing time highly depends on the size of text changes, as it requires more processing using MLMs. 

We didn't reproduce the performance with the original ORES model, as it uses caching in its API and is deployed on different instances. At the same time, the original ORES paper report median, 75\%, and 95\% percentile response timings are 1.1, 1.2, and 1.9 seconds respectively, for single score speed~\cite{halfaker2020ores}. Having those values as a reference, we can say that our solution is comparable in time performance with the ORES model but can be slightly less efficient in case of revisions with major text changes. 

% \todo{Add details on the instance that serves the model if possible}

\subsection{Fairness and biases}
\label{fairness_section}
Although anonymous edits tend to be reverted more than the ones done by registered users, the big majority of Wikipedia editions keep allowing anonymous editors (a.k.a., IP Editors).\footnote{\url{https://en.wikipedia.org/wiki/Wikipedia:Wikipedia_is_anonymous}} Therefore, it is important that ML-systems try to mitigate potential biases against this group of editors. 

This section presents a thorough analysis of the biases present in our proposed model and a comparison to existing solutions. As for analysis, we are using Disparate Impact Ratio ($DIR$) and Difference in AUC score for anonymous and registered users to analyze the bias against (unprivileged) anonymous users~\cite{aif360}. Equation \ref{eq:DIR} shows the calculation logic of $DIR$, where $Pr$ corresponds to probability, $\hat{Y}$ - predicted value, $D$ - a group of users (anonymous or registered). The results are presented in Table \ref{table_fairness}.

\begin{equation} \label{eq:DIR}
\frac{\operatorname{Pr}(\hat{Y}=1 \mid D=\text{unprivileged})}{\operatorname{Pr}(\hat{Y}=1 \mid D=\text{privileged})}
\end{equation}

We use the same unbalanced dataset as in Section \ref{final-validation} for the following experiment. Also, we calculate $DIR^{base} = 7.93$ where we use $Y$ - true values instead of $\hat{Y}$, which shows the ratio of base rates (disparate impact of the original dataset). We compare this value with the ones calculated for each algorithm. 

The first insight is that the current model (ORES) has a DIR equal to 20.02, much bigger than the base one. It means ORES presents a significant bias against the unprivileged class (anonymous users). Also, we observe the difference in performance for anonymous and registered users is -0.043. At the same time, the best-performing configuration of our solution shows a DIR value of 9.54, which is much closer to the base value. It means that we still introduce the bias against the anonymous user, but it is significantly lower. Also, the model's performance for anonymous and registered users is more similar based on the difference in AUC, which is -0.017 for the presented model. Moreover, we observe the absolute difference in AUC decrease after adding MLMs features from -0.035 to -0.017.

\begin{table}[t]
\begin{center}
\caption{Fairness metrics evaluation.}
{\tabcolsep=3pt
\begin{tabular}{p{2.5cm}|p{1.2cm}|p{2.5cm}}
\hline\label{table_fairness}
\textbf{Model} & \textbf{DIR\tablefootnote{The closer to the the $DIR^{base} = 7.93$, the better} } & \textbf{AUC difference}  \\
\hline
ORES & 20.02 & -0.043  \\
\hline
Multilingual\textsuperscript{anon}& 1.98 & 	0.073   \\
\hline
Multilingual\textsuperscript{anon} + MLM features & 2.06 & 0.084   \\
\hline
Multilingual\textsuperscript{all}  & 2.91 & \textbf{0.010}   \\
\hline
Multilingual\textsuperscript{all} + MLM features & 3.08 & 0.017  \\
\hline
Multilingual\textsuperscript{all} + user features  & \textbf{9.36} & -0.035 \\
\hline
Multilingual\textsuperscript{all} + user features + MLM features  & 9.54 & -0.017   \\
\hline
\end{tabular}%
}
\end{center}
\vspace{-15pt}
\end{table}

\subsection{Performance on different languages}

Previously, the system design implemented independent model training for each of the languages~\cite{halfaker2020ores}. In our setup, we train one model that takes the language as a feature. Such a design makes it easier to maintain and extend models. Moreover, we extend the number of supported languages from 28 to 47.

\begin{figure}[th]
  \centering
  \includegraphics[width=\linewidth]{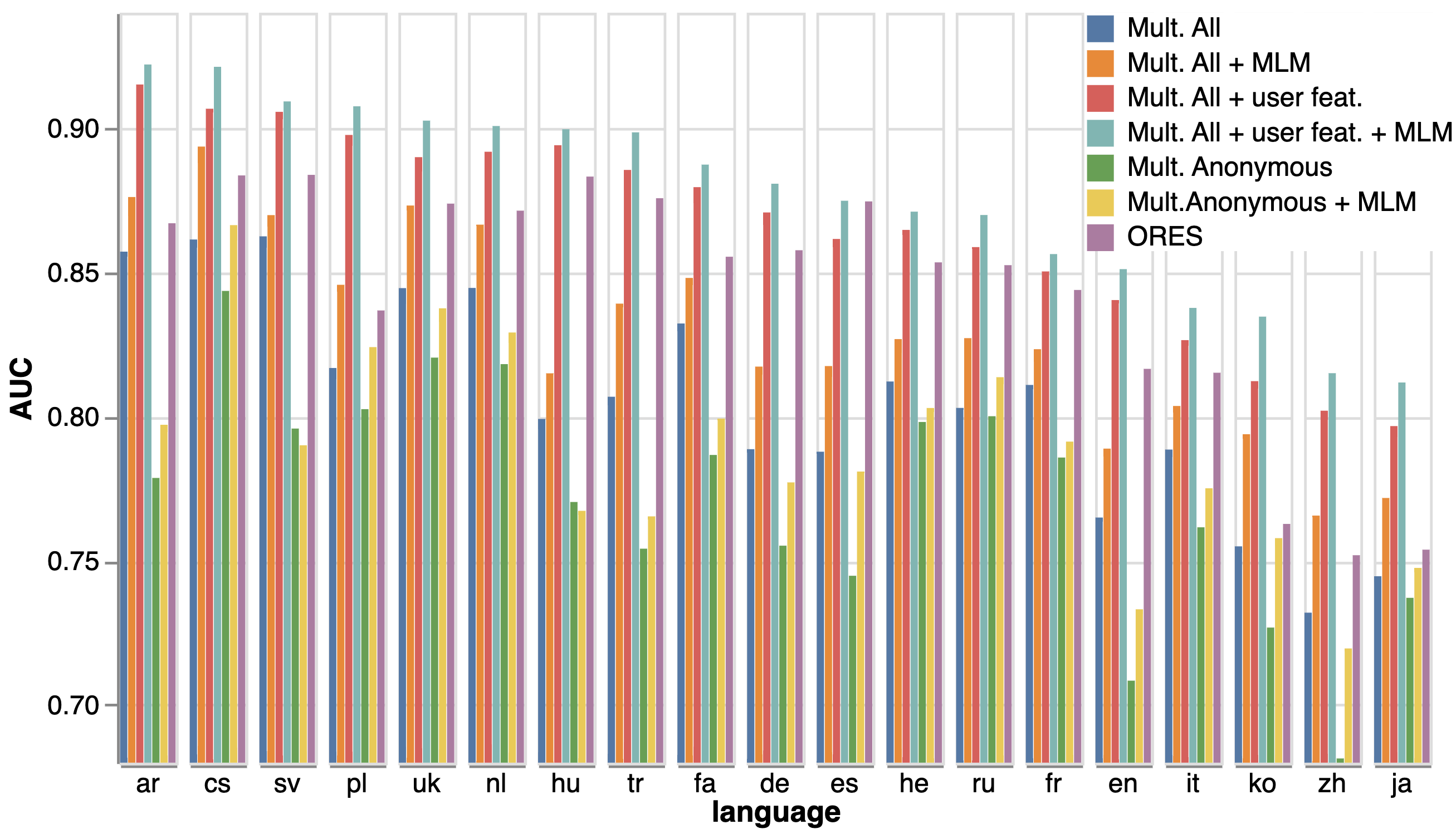}
  \caption{AUC score per model and language.}
  \label{bar_comparison}
\end{figure}

At the same time, in Section~\ref{eda_section}, we show that languages significantly differ in their characteristics like revert rate and rate of anonymous users. In this section, we observe and compare models' performance across languages supported by the ORES model and have at least 10K observations in the test dataset. We compare the models using the AUC score, and the results are presented in Figure~\ref{bar_comparison}. We not only significantly increase the language coverage, but also our solution performs better or equal to ORES for all previous languages. Moreover, we can observe that for some languages, we can achieve better performance than ORES even without user-based features that introduce bias against anonymous users. It supports the claim that our model usage can benefit each independent language community. 

Moreover, we show how the general model performance depends on the rate of anonymous users per language. We observe the negative correlation between the anonymous user rate and AUC metric both for ORES and our model. At the same time, the new model shows significantly better performance for most languages. 

% \question[author=Mykola]{
% There is interesting behavior here for the scatter-plot. The correlations are -0.69 for our model and -0.62 for ORES. However, there is an outlier - eswiki. If we remove it, the correlations will be -0.69 and -0.74, respectively. Basically, it will mean that we reduce the impact of anonymous users rate on model performance. This behavior is mostly caused by the fact that we increased performance for all languages but not for es
% }

\begin{figure}[th]
  \centering
  \includegraphics[width=\linewidth]{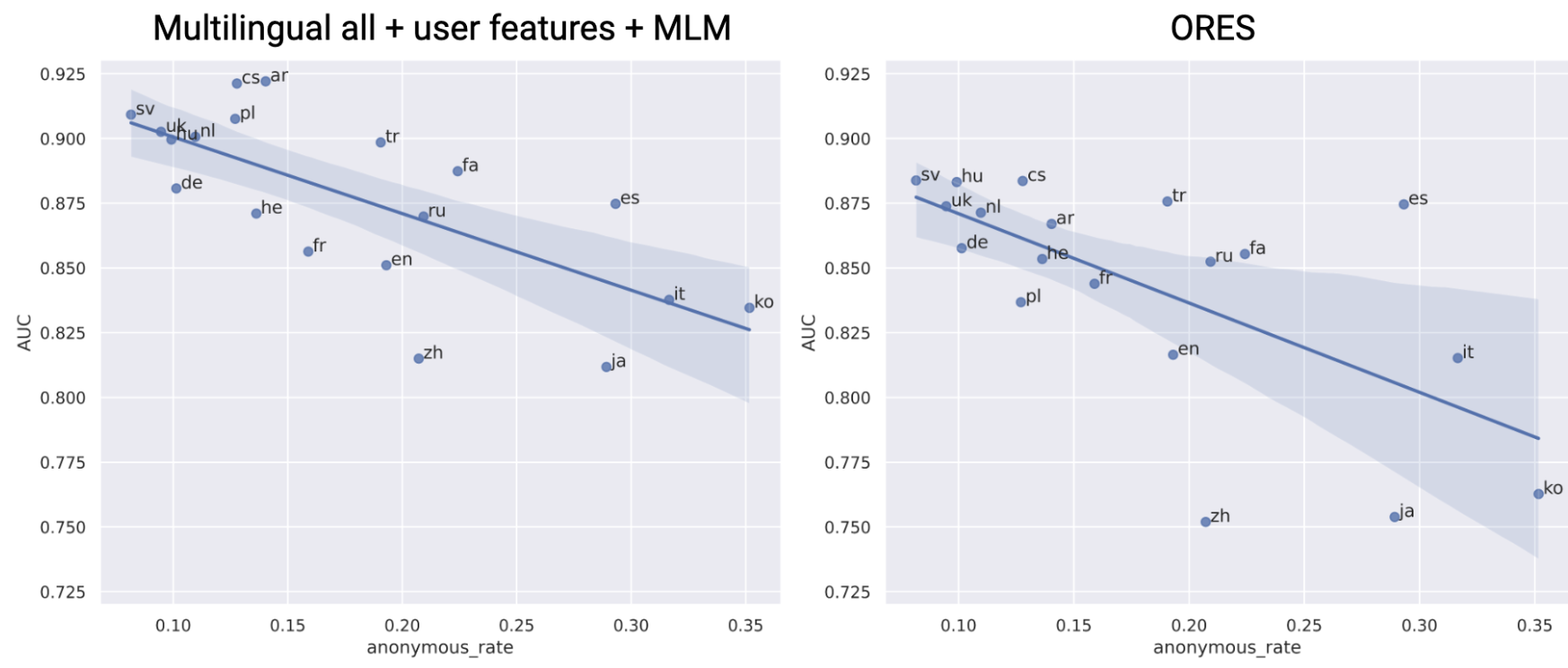}
  \caption{Anonymous rate vs. AUC per language.}
  \label{bar2}
% \vspace{-3mm}
\end{figure}

% \begin{figure}[t]
%   \centering
%   \includegraphics[width=\linewidth]{pictures/barplot2.png}
%   \caption{AUC score for Multilingual model for languages that are not available in ORES.}
%   \label{bar2}
% \end{figure}

% \question[to=Diego]{I remember you mentioned some specific usage of ORES for Spanish Wikipedia, but I can't remember the details. Can it be the reason why there is some bias in collected data giving the benefit to the ORES model on evaluation?}

\subsection{Model selection and explainability}

In recent years, there has been a growing interest in developing machine learning models that are accurate and interpretable. Explainability, or the ability to understand the reasoning behind a model's predictions, is crucial for building trust in the model and for identifying potential biases or errors. Here we focus on two key aspects of explainability: model feature importance and single-sample prediction explanation. We are analyzing the best model based on the evaluation in the previous sections and we use SHAP (SHapley Additive exPlanations) values for it.

SHAP  values calculation is a method for interpreting the output of machine learning models by attributing each feature's contribution to the model's prediction.
% They are based on the mathematical concept of Shapley values from cooperative game theory. 
SHAP values provide a way to decompose the model's prediction into a sum of the contributions of each feature. It allows the user to understand how each part is contributing to the final result~\cite{SHAP}.

\begin{figure}[tb]
  \centering
  \includegraphics[width=\linewidth]{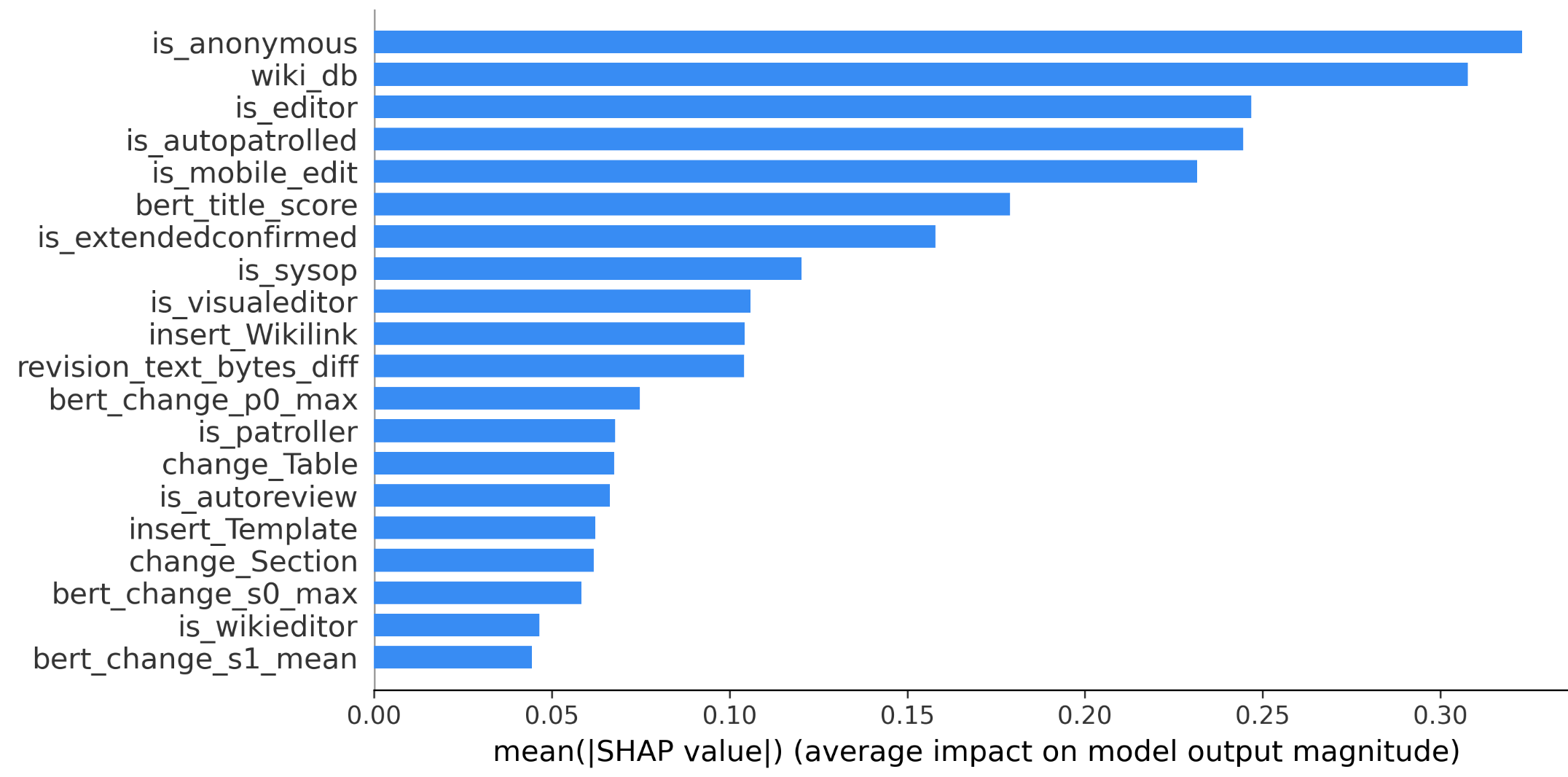}
  \caption{Top-20 feature importance based on SHAP values for all users test set.}
  \label{feature_importance}
% \vspace{-12pt}
\end{figure}

Firstly, we observe the feature importance of the new model shown in Figure~\ref{feature_importance}. We use mean absolute SHAP values for this experiment. We can see that the most influential feature is the indicator if the user is anonymous or not. This insight was not surprising, as we have shown in Section~\ref{final-validation}, that baseline, that using only this feature can achieve 0.75 in AUC, meaning that the feature has strong predictive power. On the other hand, it introduces bias against anonymous users, so we aimed to reduce it by adding other features with a different nature that also have predictive power, like MLMs features. So, we can see that 4 out of 20 most important features of the model are MLM-based when the model has 137 features in total. 

\begin{figure}[tb]
  \centering
  \includegraphics[width=\linewidth]{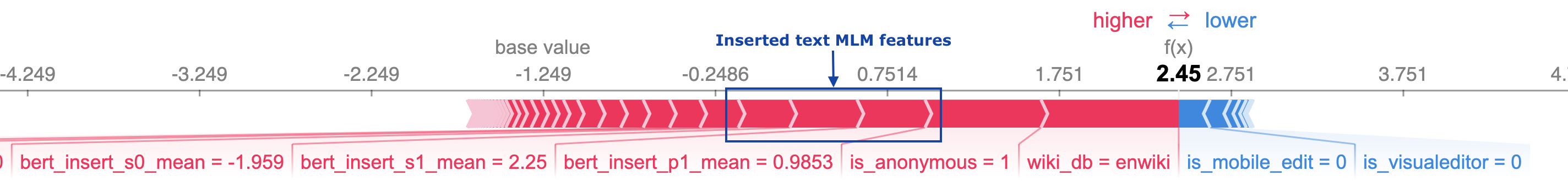}
  \caption{SHAP explanation for data sample from Figure~\ref{datasample}.}
  \label{shap_local}
% \vspace{-4mm}
\end{figure}

Moreover, we also observe a local explanation of the prediction for the sample shown before in Section~\ref{eda_section} in Figure~\ref{shap_local}. One of the most influential features for this particular example is MLMs features based on inserted text. Looking at the inserted text in Figure~\ref{datasample}, we can observe abuse text that definitely should be reverted. So, for this example the model explanation using the SHAP value provided a correct insight into the possible reason for the problem with a revision. It can define more granular problems for each revision and indicate possible problems for the patrollers and contributors. 

% \begin{figure}[t]
%   \centering
%   \includegraphics[width=\linewidth]{pictures/importance_anon.png}
%   \caption{Feature importance based on SHAP values for anonymous users test set}
%   \label{bar2}
% \end{figure}

% \todo{Add features importance for the best model from the previous section.}
% \todo{Use SHAP for analyzing: SHAP feature importance for different groups of users. For the best model <- doubtfully, as there are no valuable insights there.}
% \todo{Add SHAP per sample example for the best model from the previous section.}
% \todo{scatter plot: accuracy vs. (revert rate, number of articles, number of active users, rate_of_reverts_by experienced users)}

\section{Conclusions and Future Work}
\label{sec_conclusions}

% \todo{
% \begin{itemize}
%     \item Observe the results and contributions of the paper
%     \item Discuss the limitations
%     \item Discuss the further work
% \end{itemize}
% }

%%%%%%%%%%%%%%%%%
To sum up, this paper presents a new design for a system aimed to help in preventing vandalism on Wikipedia. The model was trained using a large dataset of 47 different languages that utilize sophisticated filtering and feature engineering techniques. 
The presented solution takes advantage of modern NLP techniques that rely on pre-trained multilingual MLMs. It also differs from the previous solution introducing a single model for all languages, making it easier to maintain. 
The results showed that the system outperformed the current state-of-the-art ORES system in terms of both the number of languages covered and accuracy. This research contributes to the field of multilingual Wikipedia patrolling, making it more efficient and less biased against anonymous users worldwide.

Improving model accuracy directly influences the encyclopedia's knowledge integrity because patrollers can cover more damaging pages with the same resources. Moreover, reducing bias against anonymous users aims to help extend the active editors pool, which is the core of the Wikipedia movement.

%\subsection{Limitations and future work}
Nevertheless, the current study has several limitations that should be considered when interpreting the results. 

\textbf{Data Limitations}: Our study focused on analyzing changes in the text content of pages. However, we have observed that only 56\% of revisions have at least some changes in the text. We are not considering changes in references, media, tables, etc. Analyzing other types of content can be a great extension of our work in the future. We restricted our data to the most frequent languages, which may not represent the full range of languages used on Wikipedia. %, which may not accurately represent the entire Wikipedia community.

\textbf{Data sampling limitations}: We make our analysis on the fixed time frame. We don't consider time-related patterns that could affect the amount of vandalism on Wikipedia and model performance. 

\textbf{Modeling limitations}: We only tested one language model and did not explore other language models that could potentially provide different results. We leave this for future work.

% Limitations: 
% \begin{enumerate}
% \item We have only ~50\% of revisions with content changes in the text. At the same time, we don't analyze the changes in references, media etc.
% \item We limit amount of data to the most frequent languages
% \item We don't consider time patterns (war etc.)
% \item Didn't tested more language models (only mbert)
% \end{enumerate}

%%
%% The acknowledgments section is defined using the "acks" environment
%% (and NOT an unnumbered section). This ensures the proper
%% identification of the section in the article metadata, and the
%% consistent spelling of the heading.
% \begin{acks}
% To Robert, for the bagels and his explanation of CMYK and color spaces.
% \end{acks}

%%
%% The next two lines define the bibliography style to be used, and
%% the bibliography file.
\begin{acks}
This work has been funded by MCIN/AEI /10.13039/501100011033 under the Maria de Maeztu Units of Excellence Programme (CEX2021-001195-M)
\end{acks}

\onecolumn
\begin{multicols}{2}
\bibliographystyle{ACM-Reference-Format}
\balance
\bibliography{main_references}
\end{multicols}

\end{document}